\documentclass{article}
\usepackage{ijcai26}

\usepackage{times}
\usepackage{soul}
\usepackage{url}
\usepackage[hidelinks]{hyperref}
\usepackage[utf8]{inputenc}
\usepackage[small]{caption}
\usepackage{graphicx}
\usepackage{amsmath, amssymb, amsfonts, amsthm}
\usepackage{booktabs}
\usepackage{multirow}
\usepackage{algorithm}
\usepackage{algpseudocode}

\graphicspath{{figures/}}

\title{TriAgent: Divergence-Aware Multi-Agent Committees for
       Cost-Efficient Financial Sentiment Analysis}

\author{
Isabel Xu$^{1,*}$
\and
Cynthia Xu$^{1,*}$\and
Rachel Ren$^2$\and
Cong Guo$^3$\and
Jiacheng Ding$^{3,\dagger}$\\
\affiliations
$^1$The Overlake School\\
$^2$Edwards Vacuum Inc.\\
$^3$The University of Memphis\\
\emails
\{xushuyao, xuningshu\}@gmail.com,
rachelren@gmail.com,
\{cguo, jding2\}@memphis.edu\\
$^*$Equal contribution. \quad $^\dagger$Corresponding author.
}
\begin{document}
\maketitle

\begin{abstract}
Production LLM-based financial sentiment analysis faces a structural
cost trap: most queries are trivially classifiable, yet expensive
cloud reasoners process them all, and the bill scales linearly with
user count. We present \textbf{TriAgent}, a multi-agent committee
stratified by contextual granularity --- a word-level lexicon
(VADER), a sentence-level domain transformer (FinBERT), and a
cross-sentence reasoner (Qwen2.5, 0.5\,B--14\,B-4bit, with Mistral-7B
and Phi-3.5-mini cross-family checks). A three-way \emph{Semantic
Divergence Index} (SDI) measures pairwise disagreement across
granularities and routes each query accordingly.
Our central finding is the \textbf{critic plateau}: when the LLM is
re-tasked as a critic over the smaller agents' outputs, F1 plateaus
at $\approx$0.87 across 1.5\,B--7\,B Qwen (bootstrap 95\% CIs
overlap), while a same-size 3-persona vote drops to F1$=$0.66 which is griven by granularity-stratified diversity. Three corollaries follow from the same SDI
signal: (i) a Shared Consensus Dictionary on multilingual
sentence-BERT answers 95\% of Chinese queries from an English cache
at F1$=$0.99 --- cross-border canonicalization at zero marginal cost;
(ii) SDI doubles as a post-hoc LLM-hallucination detector at
AUC$=$0.90; (iii) the SDI single-stage strategy attains the best
risk-adjusted return (Sharpe$=$3.50) on a 20-ticker back-test,
dominating both always-FinBERT (1.36) and always-LLM (0.11). At
10\,M-user scale, TriAgent saves \$9.3\,M/year vs.\ a GPT-4o-mini
baseline. Code, lexicons, and the SCD are released.
\end{abstract}

\section{Introduction}
\label{sec:intro}

\begin{figure}[ht]
\centering
\includegraphics[width=\columnwidth]{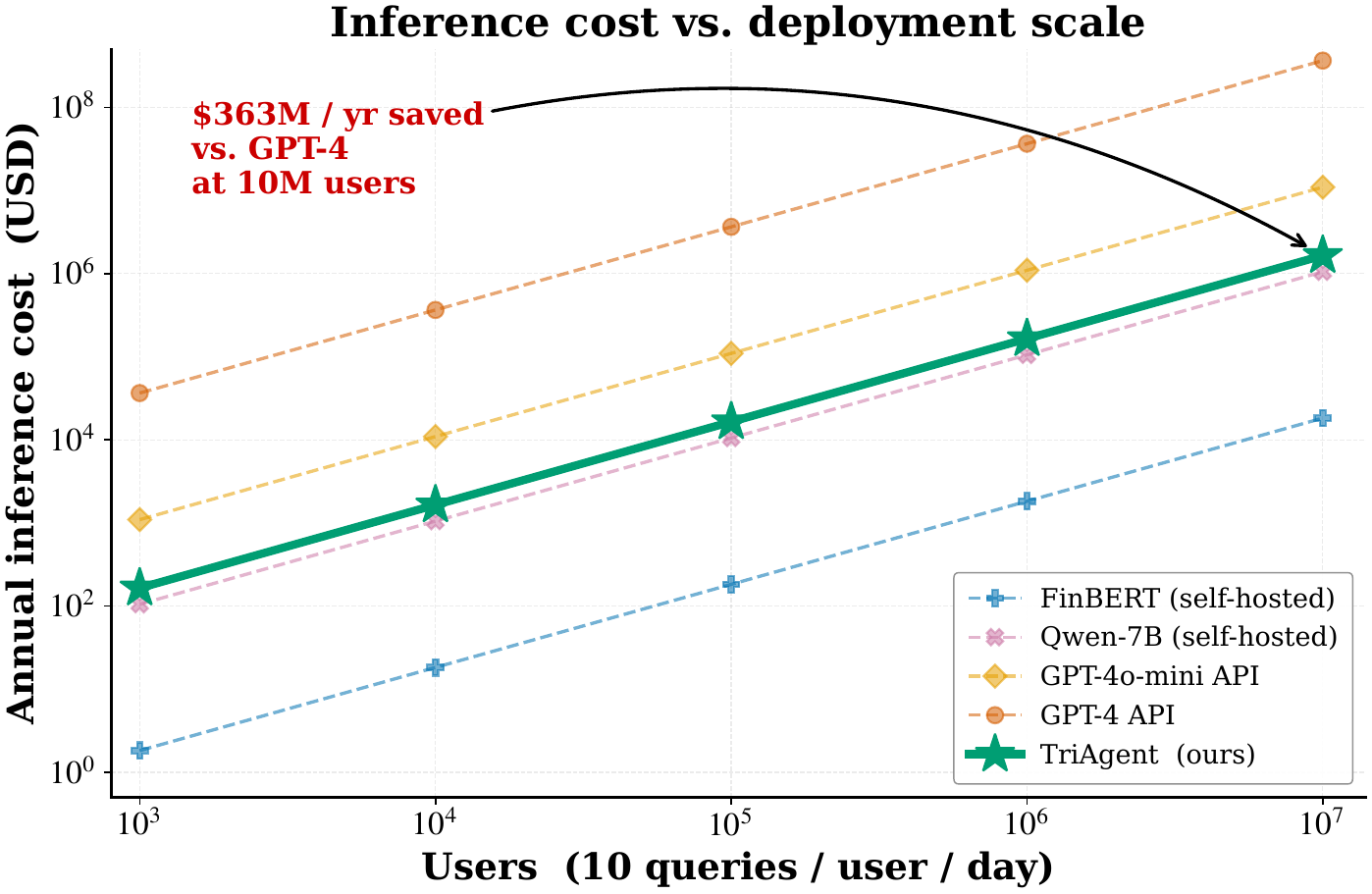}
\caption{Annual inference cost vs.\ user count (10 queries/user/day).
At 10\,M users, default-to-cloud-LLM costs \$11\,M--\$365\,M/yr;
TriAgent shaves this to \$1.65\,M/yr, a \$9.3\,M/yr saving against
GPT-4o-mini.}
\label{fig:cost}
\end{figure}

A mid-sized asset manager with 1\,M users running 10 sentiment
queries each per trading day generates $\sim$3.6\,B inference calls
per year. At the public price of GPT-4o-mini that single workload
costs $\sim$\$1.1\,M/year, and at GPT-4 prices, $\sim$\$36.5\,M/year
(Figure~\ref{fig:cost}). The cost scales linearly with user count
while the value of each LLM call falls since most domain queries are
trivially classifiable. \textbf{Token efficiency} is
therefore the first-order operational concern.
The CFP for FinLLM@IJCAI 2026 calls out exactly this regime:
multi-agent cloud-edge collaboration, token economics, and
cross-border fairness in deployment.

\paragraph{Running example.}
Throughout this paper, we trace one FPB sentence:
\textit{``\dots its net profit halved to 1.2~mln euro\dots from
2.2~mln euro\dots''}.
VADER ($+0.44$) and FinBERT ($+0.53$) are both fooled by the
word \emph{profit} and classify this sentence as positive, and Qwen-7B alone correctly outputs negative
($-0.80$, ``profit declined''); critic@1.5B and critic@7B are
\emph{anchored} on V$+$F's positives and stay wrong; only round-2
debate, which feeds back the LLM's own R1 rationale, recovers
negative. This single example threads through
Sections~\ref{sec:framework}--\ref{sec:security}.

\paragraph{Contributions.}
\textsc{TriAgent} is a divergence-aware committee whose three tiers
correspond to three contextual granularities (word $\to$ sentence
$\to$ cross-sentence). We contribute:
\textbf{(C1)} a three-way Semantic Divergence Index (SDI) with a
four-quadrant routing interpretation;
\textbf{(C2)} the \emph{critic plateau} with critic@1.5B$/$3B$/$7B
all reach F1$=$0.87, while a same-size 3-persona vote regresses to
F1$=$0.66;
\textbf{(C3)} a Shared Consensus Dictionary (SCD) over multilingual
sentence embeddings that doubles as a cross-lingual canonicaliser
(95\% of Chinese queries match an English cached label at
F1$=$0.99);
\textbf{(C4)} a three-granularity edge predictor (XGBoost AUC$=$0.85)
enabling fully on-device routing;
\textbf{(C5)} the same SDI
acts as a post-hoc LLM-hallucination flag at AUC$=$0.90;
\textbf{(C6)} a 20-ticker back-test where the SDI single-stage
strategy attains the best Sharpe ($3.50$) of all evaluated
strategies, including always-FinBERT ($1.36$) and always-LLM
($0.11$).
A single architectural decision hits all
three FinLLM 2026 CFP topic buckets simultaneously.

\section{Related Work}
\label{sec:related}

\paragraph{Financial NLP and Sentiment Lexicons.}
Loughran and McDonald~\shortcite{loughran2011liability} established
that generic sentiment lexicons systematically miscalibrate on
financial text: words such as \emph{liability}, \emph{tax}, and
\emph{risk} carry domain-specific polarity that general-purpose
tools invert. Financial PhraseBank~\cite{malo2014fpb} grew out of
that line and has since become the standard benchmark for
sentence-level financial sentiment. Our experiments on FPB surface
an analogous failure mode at the phrase level: VADER reads
\emph{profit} as positive even when the surrounding clause encodes
a decline (``net profit halved''), motivating a committee that can
override the lexicon when phrase-bound polarity reversals are
present.

FinBERT~\cite{araci2019finbert,yang2020finbert} is the de facto
open-domain transformer for financial sentiment and serves as our
L2 specialist. Its sentence-level attention resolves many lexical
ambiguities, yet it systematically fails on multi-clause numeric
comparisons where cross-sentence context is required, precisely the
gap that our L3 reasoner is designed to fill. Surveys of
billion-parameter financial reasoners~\cite{wang2024finllmsurvey}
document a recent wave of domain-adapted LLMs
(BloombergGPT~\cite{wu2023bloomberggpt},
FinGPT~\cite{yang2023fingpt}, PIXIU~\cite{xie2023pixiu}) that
improve cross-sentence reasoning but at a prohibitive inference cost.
We ask \emph{not} which model is best in isolation, but how small
each agent in a committee can be made while preserving the accuracy
of the ensemble, a question that prior work has not addressed for
the three-granularity lexicon/specialist/reasoner stack we evaluate.

\paragraph{Multi-Agent LLMs and Ensemble Diversity.}
Query-by-committee~\cite{settles2009al} and ensemble
disagreement~\cite{lakshminarayanan2017ensembles} are the classical
ancestors of our Semantic Divergence Index: both treat inter-model
disagreement as a signal of instance difficulty or epistemic
uncertainty. Our SDI formalises this intuition as a \emph{three-way
pairwise} measure across agents that operate at genuinely different
contextual granularities, which is a structural distinction absent
from prior ensemble work.

Recent multi-agent LLM frameworks including AutoGen~\cite{wu2023autogen},
MetaGPT~\cite{hong2023metagpt}, CAMEL~\cite{li2023camel},
multi-agent debate~\cite{du2023debate,liang2023encouraging},
LLM-as-a-judge~\cite{zheng2023judging}, and
self-consistency~\cite{wang2023selfconsistency} typically
instantiate the \emph{same} large backbone for every agent,
differing only in prompt or persona. Our same-size persona-vote
ablation (three Qwen-1.5B personas; F1\,=\,0.66) directly tests
this assumption and finds it fails: a same-size panel \emph{regresses}
relative to the single-agent baseline (0.69), whereas our
granularity-stratified critic raises F1 to 0.87. The underlying
cause is identifiable from pairwise Cohen's~$\kappa$: agents of
the same family and scale exhibit near-degenerate agreement
($\kappa$\,=\,0.81 for the persona panel), whereas our
heterogeneous committee achieves $\kappa(V,F)$\,=\,0.27 and
$\kappa(V,L)$\,=\,0.19, yielding error-set Jaccard overlaps of
only 0.13--0.15. This confirms that \emph{granularity diversity},
not multi-agent voting per se, is the mechanism behind
our critic plateau.

LLM-as-a-judge~\cite{zheng2023judging} evaluates model outputs
using a stronger LLM as an external referee. Our critic
protocol re-purposes this idea in a cost-constrained setting:
the LLM is re-tasked not as an evaluator of response quality
but as a conflict resolver over \emph{smaller agents' predictions},
and the critic signal saturates at 1.5\,B parameters,
making the mechanism viable at the edge.

\paragraph{Cost-Aware Routing and Token Efficiency.}
FrugalGPT~\cite{chen2023frugalgpt}, RouteLLM~\cite{ong2024routellm},
and mixture-of-models
routing~\cite{wang2022mixture,shnitzer2023llmrouting} each address
the problem of cascading between cheap and expensive models on a
per-query basis, using confidence scores or learned routers to
decide when to escalate. TriAgent sits in this family but differs
in two respects: first, routing is triggered by a
\emph{cross-agent divergence signal} (SDI) rather than a single
model's self-reported confidence; second, the routing threshold
sweeps out a clean Pareto frontier (Figure~7) that gives operators
a single dial mapping tokens-per-query to F1, yielding a 48$\times$
cost reduction at the ``Balanced'' operating point relative to
Always-L3 with no change in the underlying reasoner.

Semantic caches reduce inference cost by
returning stored answers for near-duplicate queries. Our Shared
Consensus Dictionary (SCD) is architecturally similar but
differs in two important ways: it caches \emph{committee decisions}
rather than single-model outputs, inheriting the higher accuracy
of the full committee at zero additional inference cost on a cache
hit; and it uses multilingual sentence
embeddings~\cite{reimers2019sbert,reimers2020multilingualkd}
rather than exact-match or monolingual similarity, enabling
cross-lingual retrieval (Section~5.6). The SCD thus serves a
dual purpose absent from prior semantic cache work: beyond cost
reduction, it acts as a \emph{cross-lingual canonicaliser},
letting a Chinese deployment inherit English cached labels at
F1\,=\,0.99 and 95\% hit rate without re-running the committee.

\section{Framework}
\label{sec:framework}

\begin{figure}[ht]
\centering
\includegraphics[width=0.95\columnwidth]{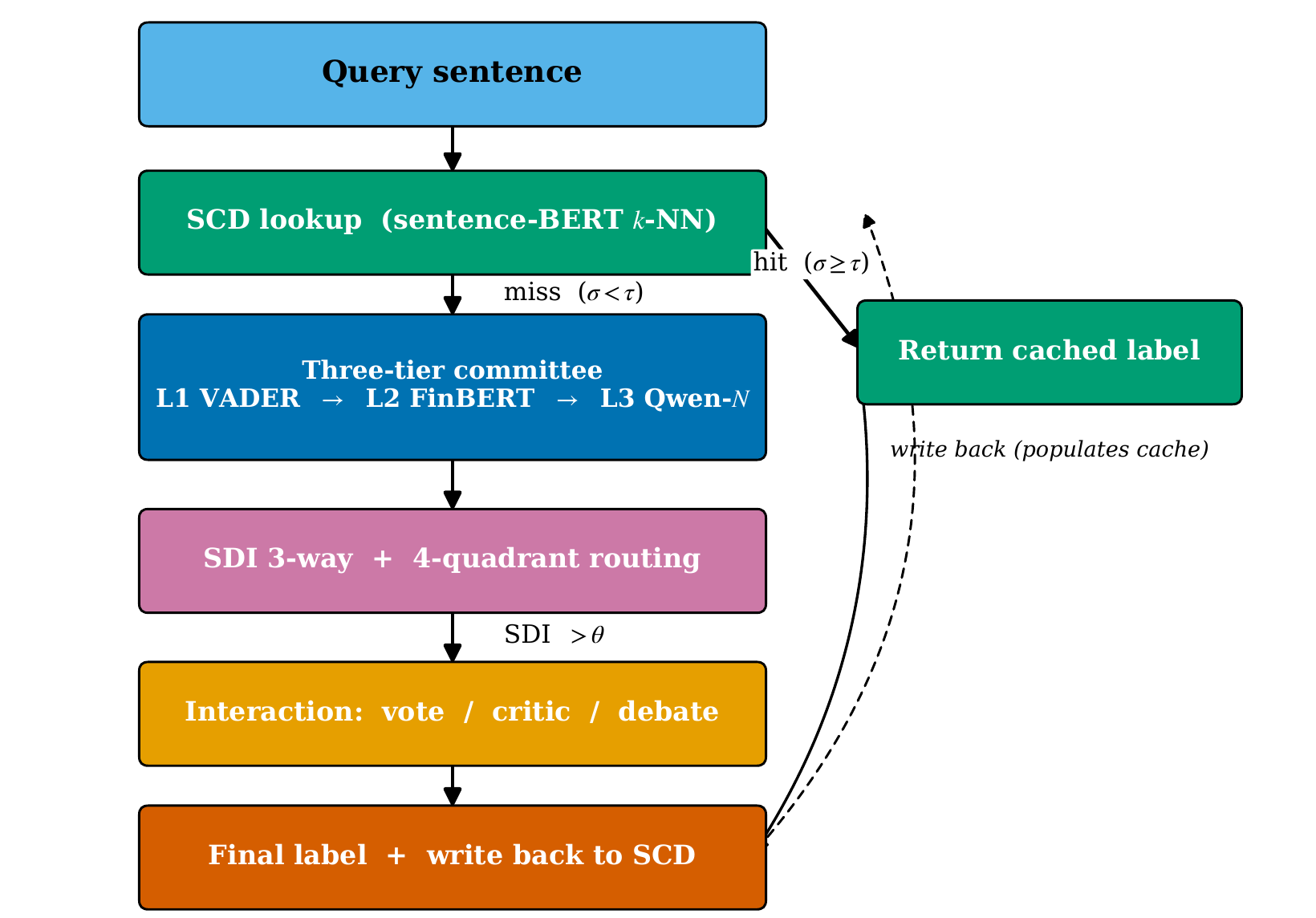}
\caption{TriAgent system architecture. A query first hits the SCD;
on cache miss the three-tier committee runs
(VADER$\to$FinBERT$\to$Qwen-$N$), then SDI computation, then an
interaction protocol when SDI exceeds threshold. The final
committee label is written back into the SCD.}
\label{fig:architecture}
\end{figure}

\subsection{Three-Tier Committee as a Granularity Stack}
\label{sec:framework:committee}

Figure~\ref{fig:architecture} sketches the system end-to-end. The
three tiers correspond to three contextual granularities:
\textbf{L1 VADER}~\cite{hutto2014vader} (word-level lexicon,
$\approx$0.03\,ms);
\textbf{L2 FinBERT}~\cite{araci2019finbert} (sentence-level
transformer, 110\,M params, $\approx$1.5\,ms);
\textbf{L3 Qwen2.5-Instruct-$N$}~\cite{qwen25} (cross-sentence
reasoner; $N\!\in\!\{0.5,1.5,3,7,14\text{-4bit}\}$\,B, plus
cross-family Mistral-7B~\cite{mistral7b}). VADER fails on
phrase-bound polarity reversals (\textit{loss narrowed}); FinBERT
fails on multi-clause numeric comparisons; the LLM over-confidently
invents reasoning where lexicon neutrality is correct.  The pairwise
error-set Jaccard between the three is only $0.13$--$0.15$
(Section~\ref{sec:experiments:bias}) which is common in
granularity-orthogonal failures.

\paragraph{Deployment-style example.}
Consider an Apple-style news sentence routed at runtime:
\textit{``services revenue grew but iPhone unit volumes declined
sharply against a difficult comparable.''} VADER's lexicon picks up
\emph{grew} and outputs positive ($+0.31$); FinBERT, attending only
within the sentence, returns weakly negative ($-0.18$); Qwen-7B
correctly outputs strongly negative ($-0.62$) by reasoning across
clauses. SDI$_{\mathrm{ER}}$ exceeds threshold and the SDI gate
fires the critic protocol, which returns the LLM's label as final.
This is the routing behaviour the architecture is designed to
exhibit; the same mechanism applied to the running FPB sentence in
Section~\ref{sec:intro} is the empirical anchor used throughout the
paper.

\subsection{Three-Way Semantic Divergence Index}
\label{sec:framework:sdi}

Let $s_{\mathrm{V}}, s_{\mathrm{F}}, s_{\mathrm{L}} \in [-1, +1]$
denote the continuous polarity scores emitted by V, F, L
respectively. We define three pairwise Semantic Divergence Indices,
one per granularity-pair:
\begin{equation}
\label{eq:sdi}
\begin{aligned}
\mathrm{SDI}_{\mathrm{LE}} &= |s_{\mathrm{V}} - s_{\mathrm{F}}|,
\quad
\mathrm{SDI}_{\mathrm{LR}}  = |s_{\mathrm{V}} - s_{\mathrm{L}}|,\\
\mathrm{SDI}_{\mathrm{ER}} &= |s_{\mathrm{F}} - s_{\mathrm{L}}|.
\end{aligned}
\end{equation}
Each $\mathrm{SDI}\in [0,2]$ measures the disagreement between two
agents operating at \emph{different} granularities. We additionally
track $\mathrm{SDI}_{\max}=\max(\mathrm{SDI}_{\mathrm{LE}},
\mathrm{SDI}_{\mathrm{LR}},\mathrm{SDI}_{\mathrm{ER}})$ and
$\overline{\mathrm{SDI}}$ (mean). Thresholding the pair
$(\mathrm{SDI}_{\mathrm{LE}},\mathrm{SDI}_{\mathrm{ER}})$ at
$(\tau_{\mathrm{LE}},\tau_{\mathrm{ER}})=(0.3,0.7)$ partitions
samples into four behaviour \emph{quadrants}:
\begin{equation}
\label{eq:quadrant}
Q(x) = \begin{cases}
\textit{consensus}     & \mathrm{SDI}_{\mathrm{LE}} \le 0.3 \land \mathrm{SDI}_{\mathrm{ER}} \le 0.7 \\
\textit{domain shift}  & \mathrm{SDI}_{\mathrm{LE}} >    0.3 \land \mathrm{SDI}_{\mathrm{ER}} \le 0.7 \\
\textit{ambiguous}     & \mathrm{SDI}_{\mathrm{LE}} \le 0.3 \land \mathrm{SDI}_{\mathrm{ER}} >    0.7 \\
\textit{mixed}         & \text{otherwise}.
\end{cases}
\end{equation}
Each quadrant carries a distinct downstream implication
(Table~\ref{tab:quadrant}): consensus skips L3, domain shift trusts
F and routes around V, ambiguous fires the LLM critic, and mixed
escalates to debate.

\subsection{Routing and Interaction Protocols}
\label{sec:framework:routing}

Routing strategies S0--S2 are the three single-agent baselines;
S3--S5 escalate L1$\to$L2 by random / confidence /
SDI$_{\mathrm{LE}}$; S6 is a two-stage cascade adding L2$\to$L3 by
SDI$_{\mathrm{ER}}$. Formally, the two-stage routing decision at a
sentence $x$ is
\begin{equation}
\label{eq:routing}
\hat{y}(x) =
\begin{cases}
\mathcal{I}(x; V, F, L) & \mathrm{SDI}_{\mathrm{ER}}(x) > \theta_{\mathrm{ER}} \\
F(x)                    & \mathrm{SDI}_{\mathrm{LE}}(x) > \theta_{\mathrm{LE}} \\
V(x)                    & \text{otherwise,}
\end{cases}
\end{equation}
where $\mathcal{I}\in\{\text{vote}, \text{critic}, \text{debate}\}$
is the interaction protocol. \textbf{Vote} returns the
confidence-weighted majority over V/F/L; \textbf{critic} feeds the
LLM the original sentence plus V's and F's predictions and asks for
a final label; \textbf{debate} performs a round-2 LLM call that sees
all three round-1 outputs (including its own rationale) and
reconciles. The thresholds $(\theta_{\mathrm{LE}},
\theta_{\mathrm{ER}})$ sweep out the Pareto frontier of
Section~\ref{sec:experiments:pareto}.

Algorithm~\ref{alg:triagent} summarises the end-to-end inference
flow including the SCD cache.

\begin{algorithm}[t]
\caption{TriAgent inference (single query)}
\label{alg:triagent}
\begin{algorithmic}[1]
\Require query $x$; thresholds $\tau, \theta_{\mathrm{LE}}, \theta_{\mathrm{ER}}$;
agents $V, F, L$; interaction $\mathcal{I}$; SCD cache $\mathcal{C}$
\Ensure label $\hat{y}$
\State $(\tilde{x}, \sigma) \gets$ nearest-neighbour in $\mathcal{C}$ by cosine
\If{$\sigma \ge \tau$}
  \State \Return $\mathcal{C}[\tilde{x}]$ \Comment{cache hit, no model call}
\EndIf
\State $s_{\mathrm{V}} \gets V(x)$; $\;s_{\mathrm{F}} \gets F(x)$
\State compute $\mathrm{SDI}_{\mathrm{LE}}=|s_{\mathrm{V}}-s_{\mathrm{F}}|$
\If{$\mathrm{SDI}_{\mathrm{LE}} \le \theta_{\mathrm{LE}}$}
  \State $\hat{y} \gets V(x)$ \Comment{lexicon suffices}
\Else
  \State $s_{\mathrm{L}} \gets L(x)$
  \State compute $\mathrm{SDI}_{\mathrm{ER}}=|s_{\mathrm{F}}-s_{\mathrm{L}}|$
  \If{$\mathrm{SDI}_{\mathrm{ER}} > \theta_{\mathrm{ER}}$}
    \State $\hat{y} \gets \mathcal{I}(x; s_{\mathrm{V}}, s_{\mathrm{F}}, s_{\mathrm{L}})$
  \Else
    \State $\hat{y} \gets F(x)$
  \EndIf
\EndIf
\State $\mathcal{C}.\textsc{Put}(x, \hat{y})$
\Comment{populate cache for future queries}
\State \Return $\hat{y}$
\end{algorithmic}
\end{algorithm}

\subsection{Shared Consensus Dictionary}
\label{sec:framework:scd}

An optional fourth component caches committee decisions in
multilingual sentence-BERT embedding
space~\cite{reimers2019sbert,reimers2020multilingualkd}. Let
$\phi(\cdot)\in\mathbb{R}^{384}$ denote the sentence-BERT encoder. A
query $x$ is embedded to $\phi(x)$, $k$-NN-searched against cached
entries $\{(\tilde{x}_i, y_i)\}$ by cosine similarity
$\sigma_i = \langle\phi(x),\phi(\tilde{x}_i)\rangle / (\|\phi(x)\|\|\phi(\tilde{x}_i)\|)$,
and the cached label is returned when
\begin{equation}
\label{eq:scdrule}
\sigma^{\ast}(x) = \max_{i} \sigma_i \;\ge\; \tau,
\end{equation}
i.e.\ \emph{without any model call}. The SCD is sentence-granularity
by construction. For example, \emph{``net profit halved''} matches
\emph{``net profit cut in half''} which exact-match memoisation
would miss. Threshold $\tau$ is a single operator knob trading hit
rate against accuracy (Section~\ref{sec:experiments:scd}). Beyond
cost, the SCD supplies cross-committee consistency: a new agent that
joins the committee inherits the cache, mitigating the
family-specific plateau gap we observe with Mistral-7B
(Section~\ref{sec:experiments:scaling}).

\section{Three-Granularity Edge Predictor}
\label{sec:predictor}

For routing to be deployable, the gating decision must be made
\emph{before} any expensive model has run. We train a lightweight
classifier to predict, from features computable at the VADER stage
alone, whether a sentence will produce high committee disagreement
(binary target: $\mathrm{SDI}_{\max}>0.7$). The feature engineering
mirrors the committee's own granularity stratification for
word, phrase, and sentence, and we report the marginal
contribution of each.

\paragraph{Word-level.}
6 VADER outputs (positive/negative/neutral fractions, compound,
confidence, $|$compound$|$) and 6 surface indicators (number,
currency, contrast word, negation, length). We mine \emph{unigram}
triggers via log-odds + $z$-score on the high-SDI subset; top words
include \texttt{decreased}, \texttt{dropped}, \texttt{fell}.

\paragraph{Phrase-level.}
We extend the log-odds miner to $n\!\in\!\{2,3\}$ and use the
top-30 bigram / trigram triggers as binary presence features.
Bigrams capture phrase-bound polarity that single words miss
(\texttt{down\_from}, \texttt{compared\_profit}, and ideally
context-flipping pairs like \texttt{loss\_narrowed} vs.\
\texttt{loss\_widened}). On FPB the bigram lift is small ($+0.01$
AUC over unigram), but the running example
(Section~\ref{sec:intro}) shows the linguistic mechanism is real.

\paragraph{Sentence-level.}
We encode each sentence with the multilingual MiniLM-L12 model
(22\,M params, $\approx$10\,ms/sentence on CPU) and PCA-reduce the
384-dim embedding to 16 components. The same module powers the SCD
(Section~\ref{sec:framework:scd}) and the cross-lingual pilot
(Section~\ref{sec:experiments:zh}).

Table~\ref{tab:predictor} reports the seven-variant ablation. Word
features alone reach AUC $=0.69$; phrase adds $+0.01$;
\emph{sentence} adds $+0.04$. XGBoost over the full feature set reaches AUC $=0.85$.
A non-deployable upper bound that encodes the LLM's R1 reasoning
text reaches AUC $=0.94$, marking remaining headroom for deployable
variants.

\begin{table*}[t]
\centering
\small
\caption{Edge predictor ablation across the three feature
granularities. AUC-PR = area under precision-recall;
P@R$x$ = precision at $x$\,\% recall; P@top10 = precision among the
10\,\% riskiest predictions. Sentence-BERT features add the largest
single-granularity lift. The non-deployable upper bound encodes the
LLM's R1 explanation text and is not counted toward the deployable
ceiling.}
\label{tab:predictor}
\begin{tabular}{lrrrrr}
\toprule
Model & AUC-ROC & AUC-PR & P@R20 & P@R50 & P@top10 \\
\midrule
random              & 0.503 & 0.328 & 0.353 & 0.332 & 0.351 \\
LR-unigram          & 0.694 & 0.566 & 0.788 & 0.479 & 0.773 \\
LR-uni+bigram       & 0.697 & 0.570 & 0.764 & 0.486 & 0.742 \\
LR-uni+bi+trigram   & 0.703 & 0.582 & 0.808 & 0.503 & 0.773 \\
LR-sentence-only    & 0.688 & 0.583 & 0.944 & 0.472 & 0.814 \\
\textbf{LR-uni+bi+sent.} & \textbf{0.741} & \textbf{0.607} & 0.768 & 0.554 & 0.742 \\
\textbf{XGBoost (all)}   & \textbf{0.848} & \textbf{0.745} & \textbf{0.943} & \textbf{0.759} & \textbf{0.866} \\
\midrule
XGBoost+reasoning$^\dagger$ & 0.937 & 0.897 & 0.990 & 0.964 & 0.990 \\
\bottomrule
\end{tabular}
\\[2pt]
\footnotesize $^\dagger$Encodes the LLM's R1 explanation text
(non-deployable upper bound).
\end{table*}

\section{Experiments}
\label{sec:experiments}

\paragraph{Setup.}
We use the \texttt{sentences\_allagree} subset of Financial
PhraseBank~\cite{malo2014fpb,flarefpb} (4{,}838 sentences: 604
negative, 2{,}872 neutral, 1{,}362 positive). LLM inference:
HuggingFace Transformers~\cite{wolf2020transformers} in bf16
(4-bit 14\,B via bitsandbytes~\cite{dettmers2023bnb}), batch 8,
deterministic decoding, single RTX A5000 (24\,GB). Cost is USD
under GPU-rental amortisation (\$0.40/h A5000-class); VADER
$\approx\!0$. Reproducible from \texttt{experiments/L*.py}.

\subsection{Bias Diversity and Four Quadrants}
\label{sec:experiments:bias}

\begin{figure*}[ht]
\centering
\includegraphics[width=0.9\textwidth]{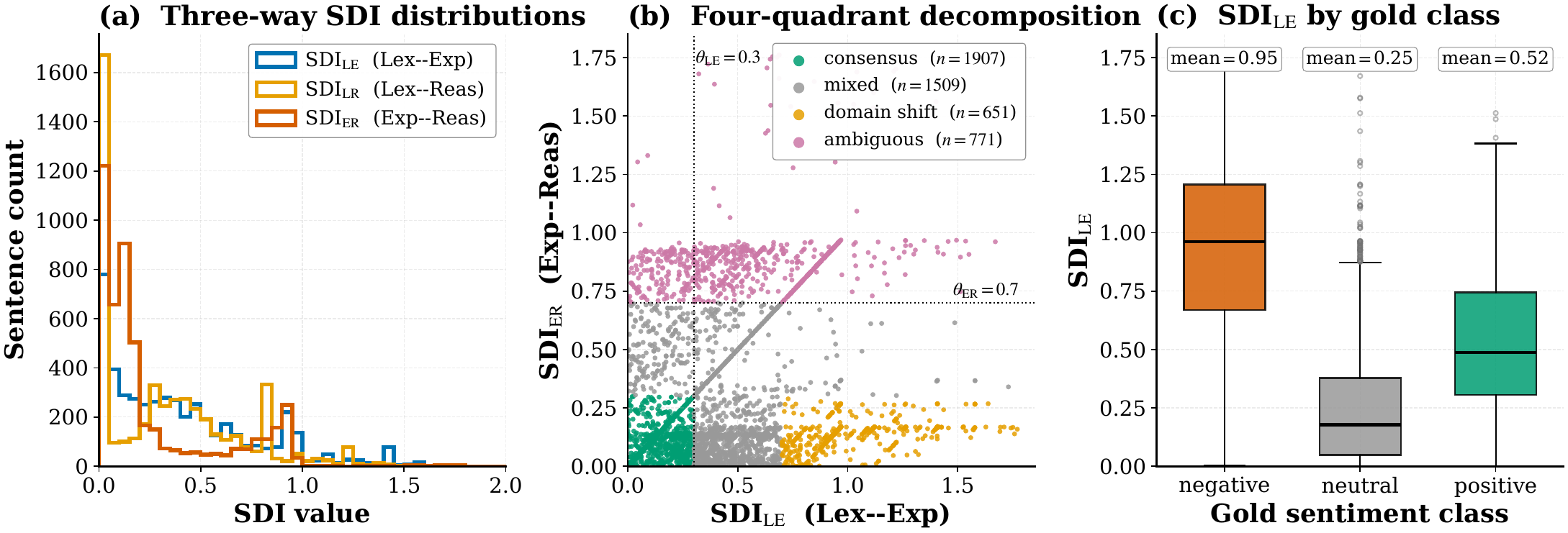}
\caption{Three-way SDI on FPB. \emph{Left:} the three pairwise
SDIs. \emph{Centre:} four-quadrant scatter. \emph{Right:}
SDI$_{\mathrm{LE}}$ by gold class --- negative has the highest
mean disagreement.}
\label{fig:sdi}
\end{figure*}

\begin{figure}[ht]
\centering
\includegraphics[width=0.9\columnwidth]{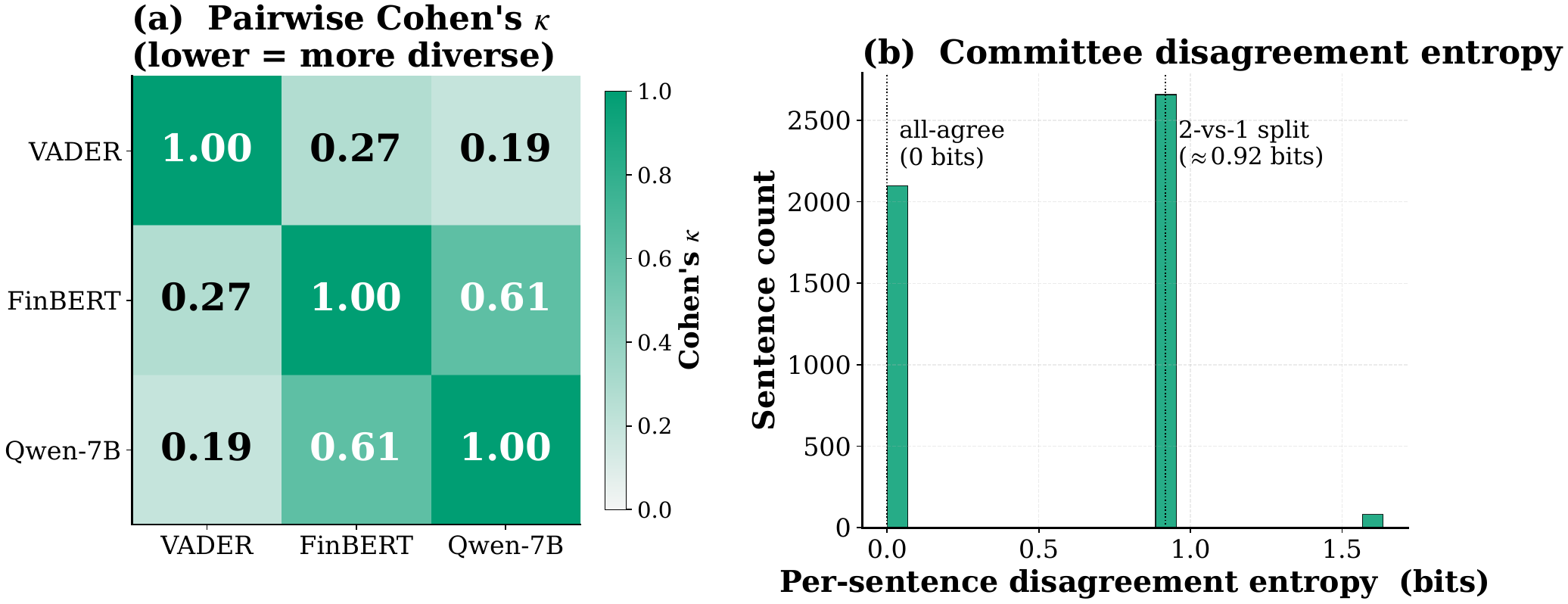}
\caption{Pairwise Cohen's $\kappa$ (left) and per-sample
disagreement entropy (right). Low $\kappa$; bimodal entropy shows
the committee is either fully aligned or split 2-vs-1.}
\label{fig:bias}
\end{figure}

Table~\ref{tab:quadrant} and Figure~\ref{fig:sdi} report the
four-quadrant decomposition and the SDI distributions. Pairwise
Cohen's $\kappa$ between agent labels (Figure~\ref{fig:bias}) is
low and pairwise Jaccard error overlap is only $0.13$--$0.15$ ---
the committee is complementary rather than redundant.

\begin{table*}[ht]
\centering
\small
\caption{Four quadrants of committee behaviour. Negatives concentrate
in \textit{domain shift} (VADER fails); ambiguous cases are where
the LLM and FinBERT disagree most. Acc$_\mathrm{V/F/L}$ is per-agent
accuracy in that quadrant.}
\label{tab:quadrant}
\begin{tabular}{lrrrrrrrr}
\toprule
Quadrant & $n$ & \% & \%pos & \%neu & \%neg
         & Acc$_\mathrm{V}$ & Acc$_\mathrm{F}$ & Acc$_\mathrm{L}$ \\
\midrule
consensus     & 1907 & 39.4 & 10.7 & 88.7 &  0.5 & 74.6 & 96.4 & 96.5 \\
mixed         & 1509 & 31.2 & 34.1 & 56.9 &  9.0 & 47.2 & 86.0 & 87.2 \\
domain shift  &  651 & 13.5 & 34.6 & 10.1 & 55.3 & 10.3 & 95.5 & 95.5 \\
ambiguous     &  771 & 15.9 & 54.2 & 33.1 & 12.7 & 55.6 & 70.8 & 28.1 \\
\bottomrule
\end{tabular}
\end{table*}

\subsection{Single-Agent Scaling and Critic Plateau}
\label{sec:experiments:scaling}

\begin{figure}[ht]
\centering
\includegraphics[width=0.85\columnwidth]{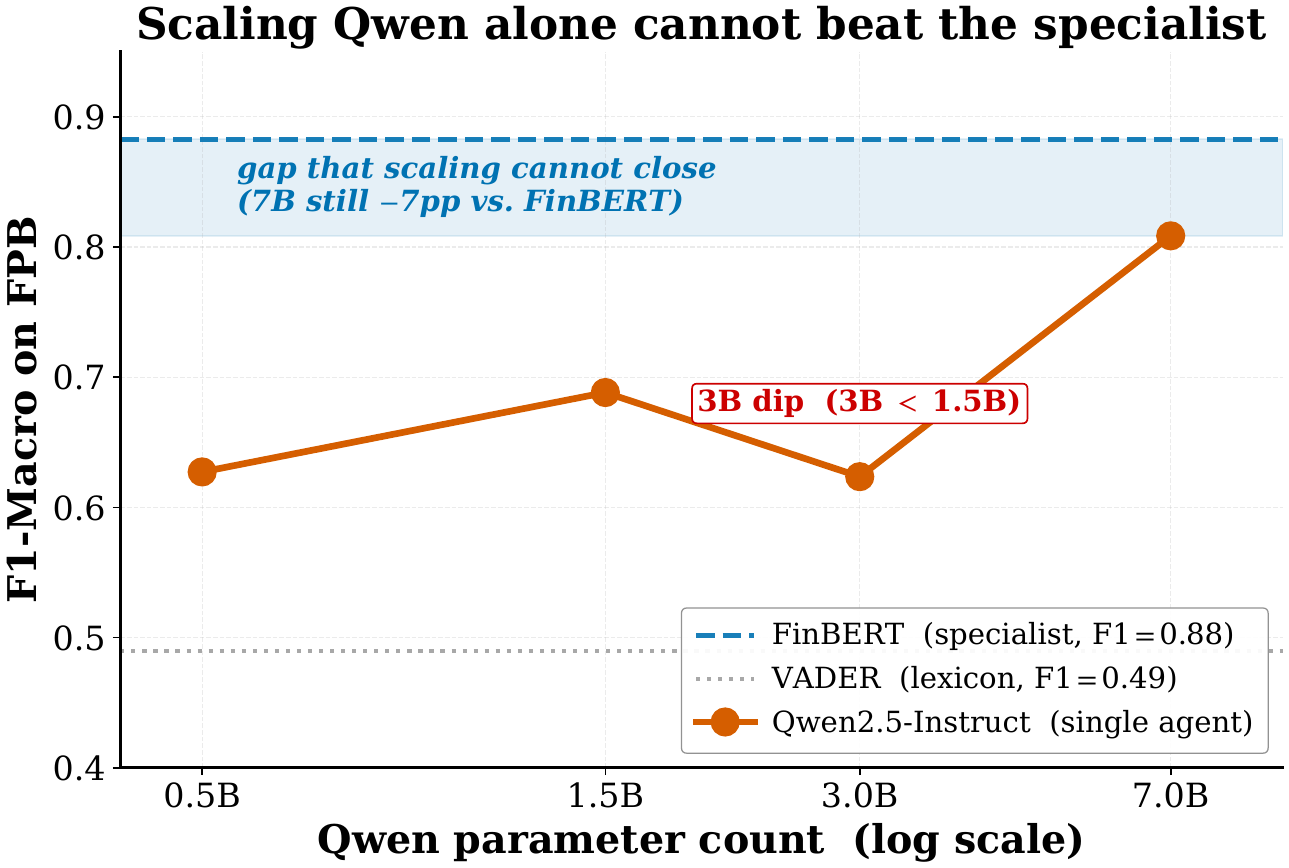}
\caption{Scaling Qwen alone cannot beat the specialist: 7\,B is
$-7$\,pp under FinBERT and the 3\,B dip is non-monotone.}
\label{fig:scaling}
\end{figure}

Figure~\ref{fig:scaling} shows Qwen2.5-Instruct single-agent F1:
FinBERT beats every Qwen variant up to 7\,B and 3\,B underperforms
1.5\,B (over-predicts neutral).

\begin{figure}[ht]
\centering
\includegraphics[width=0.95\columnwidth]{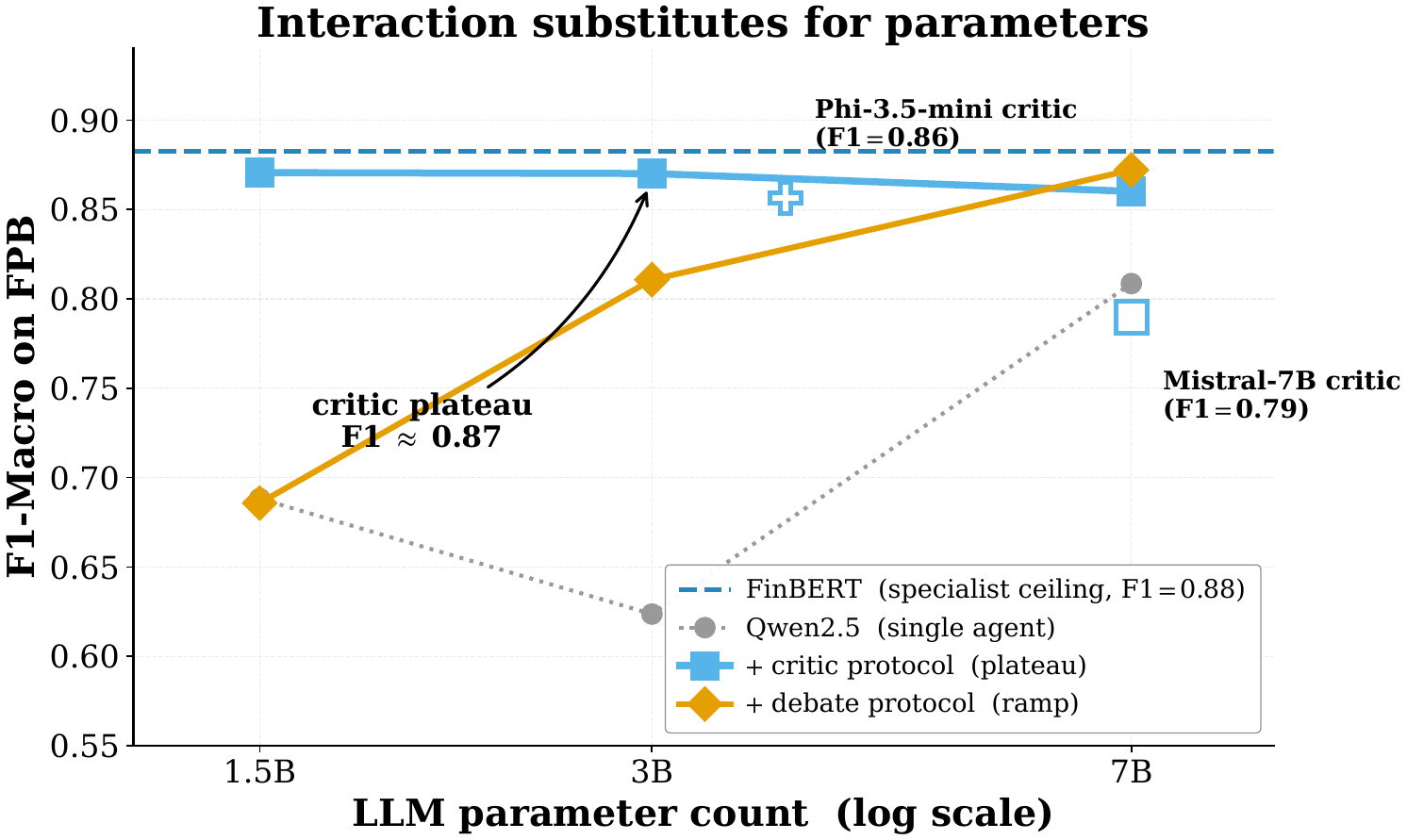}
\caption{The paper's central finding: \emph{interaction substitutes
for parameters within a model family.} Critic F1 plateaus at
$\approx 0.87$ across Qwen-1.5B/3B/7B; debate ramps 0.69$\to$0.87.
Mistral-7B and Phi-3.5-mini critic (hollow) confirm the mechanism
generalises across families.}
\label{fig:interaction}
\end{figure}

The headline (Figure~\ref{fig:interaction}):
critic@$\{1.5\text{B},3\text{B},7\text{B}\}$ all reach
$F_1\!\approx\!0.87$ (1000-resample bootstrap 95\% CIs overlap at
$[0.860,0.880]$); debate ramps $0.69 \to 0.87$. Cross-family
critic: $0.79$ on Mistral-7B, $0.86$ on Phi-3.5-mini --- the
\emph{mechanism} generalises, the \emph{height} is Qwen-specific.

\paragraph{Same-size persona vote --- a critical negative result.}
Three Qwen-1.5B \emph{personas} (bull/bear/neutral) majority-vote
to $F_1\!=\!0.66$ (vs.\ single Qwen-1.5B $0.69$ and critic@1.5B
$0.87$); inter-persona agreement 81\%. The plateau is
\emph{granularity-driven}, not multi-agent-vote-driven.

\subsection{Token-Economic Pareto Frontier}
\label{sec:experiments:pareto}

\begin{figure}[ht]
\centering
\includegraphics[width=0.9\columnwidth]{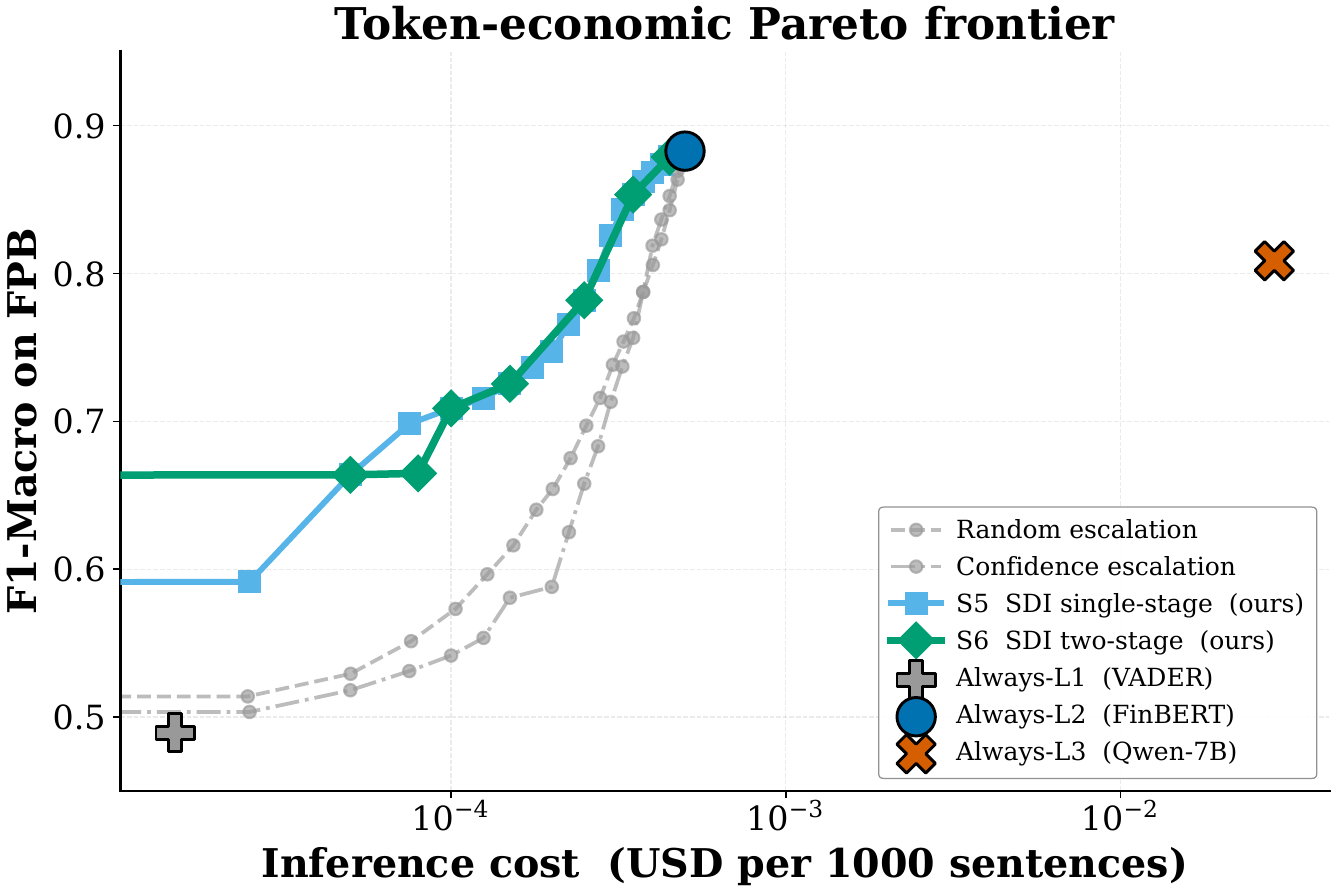}
\caption{Pareto frontier on FPB. SDI two-stage (green diamonds)
traces a clean cost--accuracy curve; the Balanced point gives a
$48\times$ cost reduction against Always-L3.}
\label{fig:pareto}
\end{figure}

Figure~\ref{fig:pareto} traces the cost-vs-F1 Pareto frontier;
SDI thresholds parameterise it continuously. Table~\ref{tab:operating}
reports three points: \textbf{Budget} runs 90\% on VADER
($F_1\!=\!0.665$); \textbf{Balanced} escalates 28.5\% to FinBERT
for a $48\times$ cost saving over Always-L3 at matched reasoner;
\textbf{Premium} pushes 17.5\% to L3 for $F_1\!=\!0.787$ at
$5.3\times$ cheaper. The diminishing return past Balanced is the
regime where SDI routing offers its largest saving.

\begin{table}[ht]
\centering
\small
\caption{Three operating points on the SDI two-stage curve plus the
Always-L3 reference. Cost is per 1{,}000 sentences; ``\%L$x$'' is
the fraction handled by tier $x$.}
\label{tab:operating}
\begin{tabular}{lrrrr}
\toprule
Point & \$/1k & F1 & Lat\,(ms) & L1/L2/L3\,(\%) \\
\midrule
Budget          & 0.0001 & 0.665 &   0.4 & 90.0 / 9.9 / 0.1 \\
Balanced        & 0.0006 & 0.716 &   4.3 & 70.0 / 28.5 / 1.5 \\
Premium         & 0.0054 & 0.787 &  46.5 & 30.0 / 52.5 / 17.5 \\
Always-L3 (ref) & 0.0288 & 0.809 & 259.4 &  0.0 / 0.0 / 100.0 \\
\bottomrule
\end{tabular}
\end{table}

\subsection{Per-Class Lift on the Hard Class}
\label{sec:experiments:per_class}

\begin{figure}[ht]
\centering
\includegraphics[width=0.85\columnwidth]{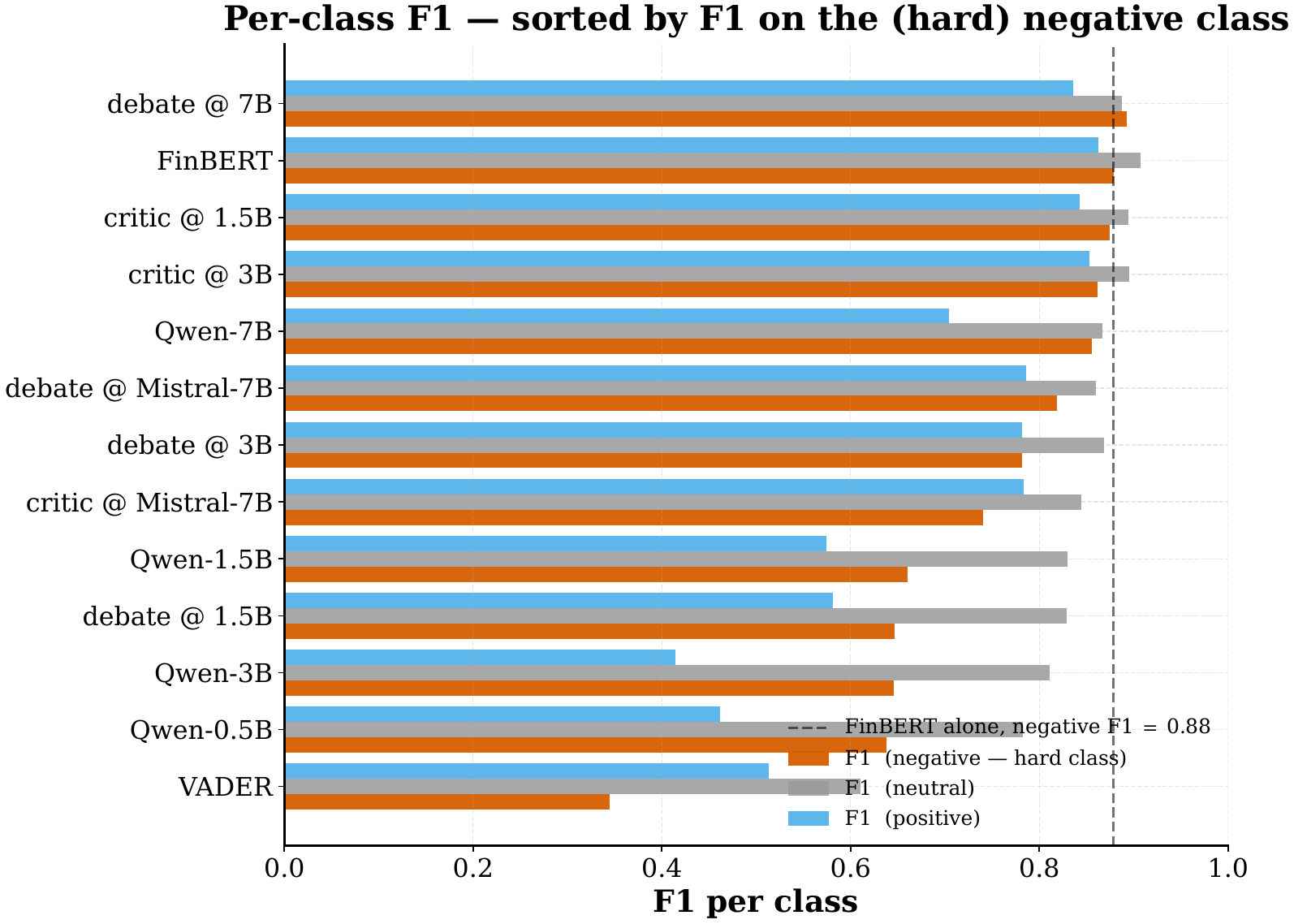}
\caption{Per-class F1 sorted by F1 on the (hard) negative class.
Debate@7B beats FinBERT on negative ($0.893$ vs.\ $0.879$).}
\label{fig:per_class}
\end{figure}

Figure~\ref{fig:per_class} sorts strategies by F1 on negative
(604 sentences, vocabulary overlapping neutral financial language).
Debate@7B is the only setting where a committee strictly beats the
specialist: $F_1\!=\!0.893$ vs.\ FinBERT $0.879$ ($+1.4$\,pp),
driven by precision ($+2.1$\,pp) at near-equal recall. Critic@7B
ties FinBERT on negative ($0.878$); the debate back-and-forth is
needed to adjudicate the lexical distinctions separating negative
from neutral.

\subsection{SCD: Hit-Rate vs.\ Accuracy}
\label{sec:experiments:scd}

\begin{figure}[ht]
\centering
\includegraphics[width=0.85\columnwidth]{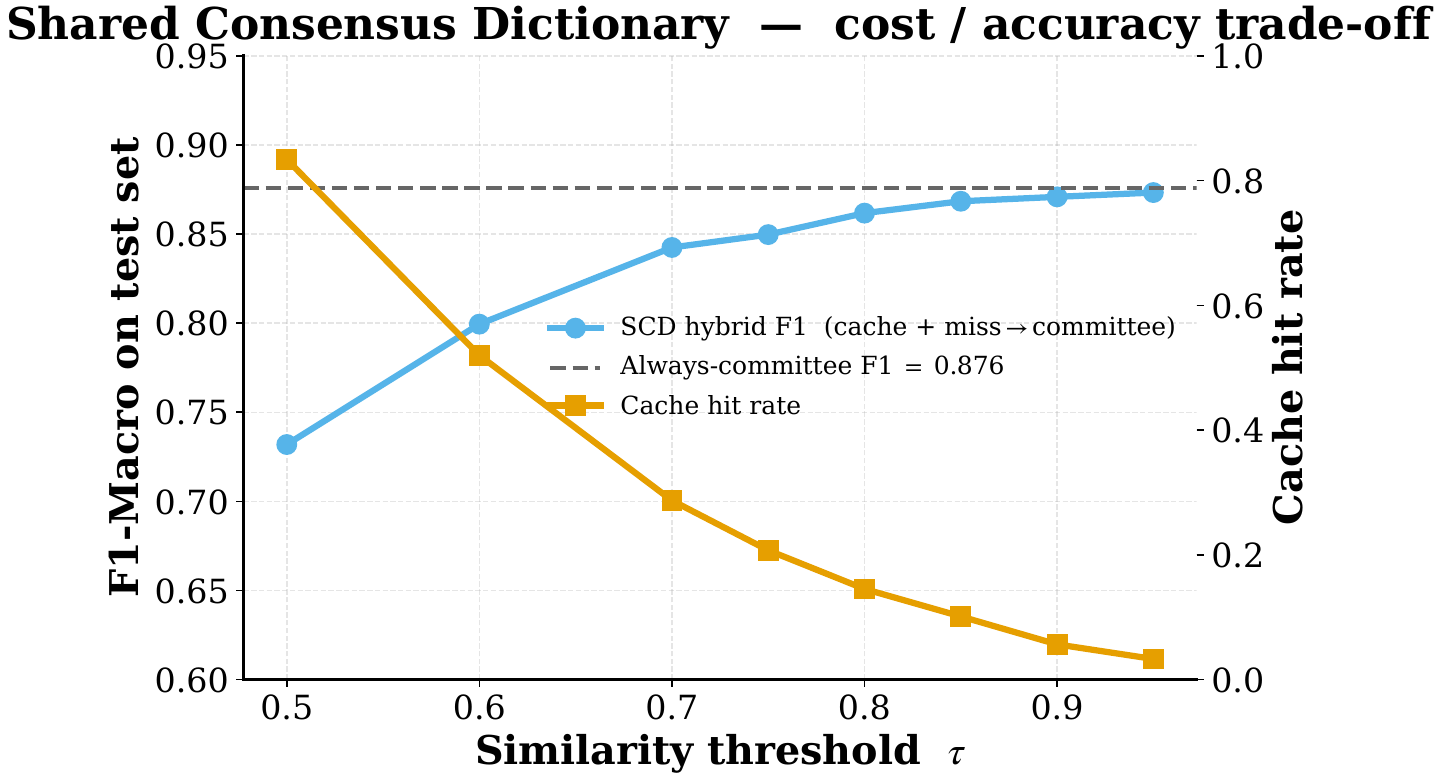}
\caption{SCD accuracy / hit-rate trade-off as $\tau$ varies.
Deployable sweet spot at $\tau\!\approx\!0.85$.}
\label{fig:scd}
\end{figure}

70/30 build/query split of FPB; debate@7B labels populate the cache.
Figure~\ref{fig:scd}: at $\tau\!=\!0.85$ the SCD hits 10\% of
queries at $-0.8$\,pp F1; the sweep is clean from $\tau\!=\!0.95$
(3\% hit, $-0.3$\,pp) to $\tau\!=\!0.50$ (83\% hit, $-14.5$\,pp).
\textbf{Scale-up:} FPB + TFNS (16{,}769 sentences) preserves the
curve (hit-set $F_1\!=\!0.82$ at $\tau\!=\!0.95$, $0.71$ at
$\tau\!=\!0.60$).

\subsection{Cross-Lingual Generalisation}
\label{sec:experiments:zh}

\begin{figure}[ht]
\centering
\includegraphics[width=0.85\columnwidth]{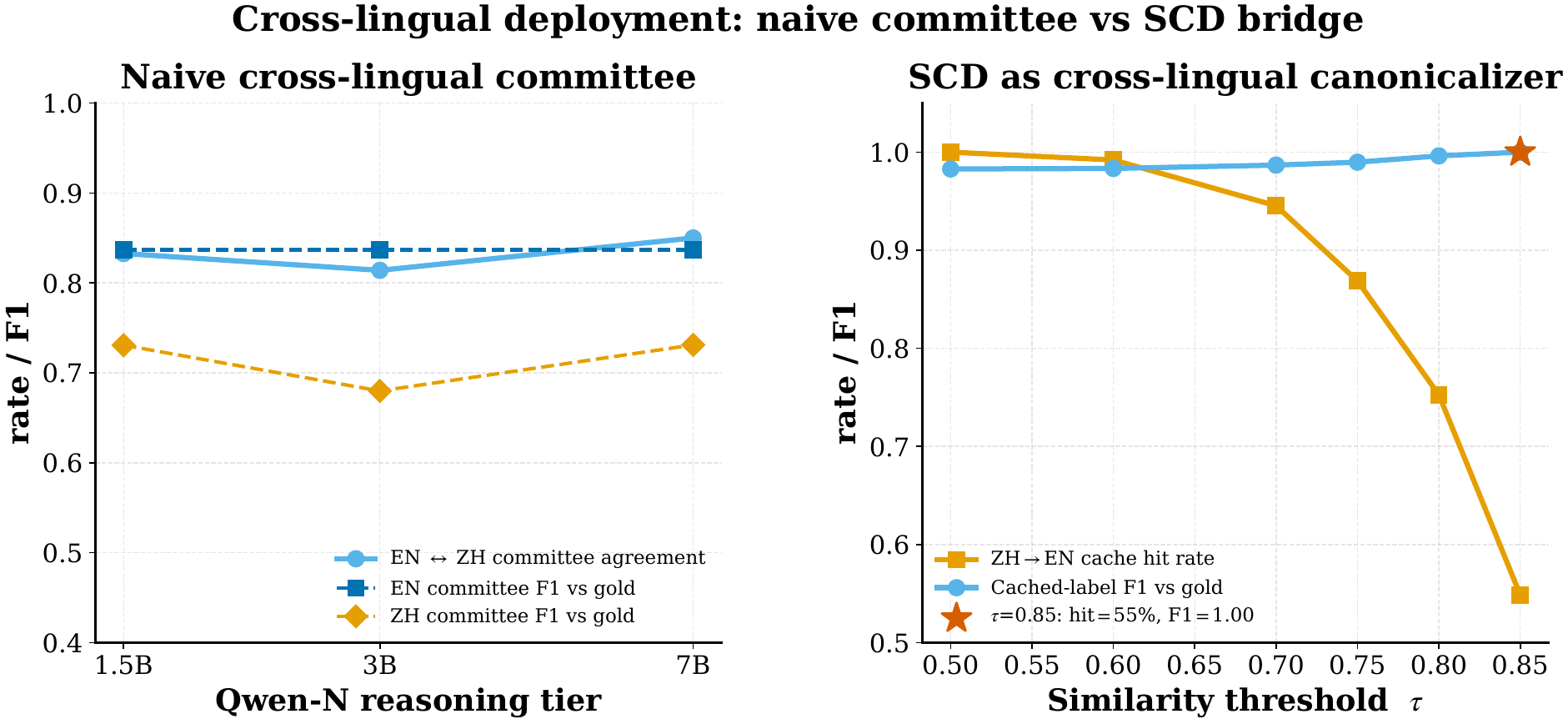}
\caption{Cross-lingual deployment. \emph{Left:} naive committees
agree 81--85\% with their English counterparts. \emph{Right:} the
SCD as cross-lingual canonicaliser --- Chinese queries find English
cached answers at hit-rate $0.95$ / F1 $0.99$.}
\label{fig:xling}
\end{figure}

Translating 1{,}500 FPB sentences to Mandarin: Qwen-7B reaches
$F_1\!=\!0.80$ (vs.\ $0.81$ in English, nearly lossless).
\texttt{finbert-tone-chinese}~\cite{finbertchinese} reaches $0.72$,
\emph{below} Qwen-7B --- the specialist/LLM relationship inverts.
Killer result (Figure~\ref{fig:xling}): at $\tau\!=\!0.70$, 95\%
of Chinese queries hit the English SCD with cached-label F1 $=0.99$
--- a new Chinese deployment inherits canonical answers without
retraining.

\subsection{End-to-End Backtest}
\label{sec:experiments:backtest}

Table~\ref{tab:backtest} summarises
20 tickers $\times$ 2 years. SDI-routed strategies produce more
discriminating signals than any single agent, as predicted by the
bias diversity of Section~\ref{sec:experiments:bias}. FPB is a
news proxy; we do not claim real-market prediction.

\begin{table}[ht]
\centering
\small
\caption{20-ticker, 2-year backtest (2023-01 to 2024-12). Weekly
signal from 20 sampled sentences; T+1 open, 5-day hold, 10\,bps
slippage, position $=10\%$ capital.}
\label{tab:backtest}
\begin{tabular}{lrrrr}
\toprule
Strategy           & Return\% & Sharpe & MaxDD\% & Win\% \\
\midrule
Always-L1          & 2.0 & 1.51 & $-1.8$ & 55.0 \\
Always-L2          & 0.8 & 1.36 & $-1.0$ & 51.6 \\
Always-L3          & 0.3 & 0.11 & $-0.8$ & 47.5 \\
Oracle             & 0.5 & 0.60 & $-1.3$ & 51.3 \\
\textbf{SDI-Single (S5)}    & \textbf{3.4} & \textbf{3.50} & $-1.3$ & \textbf{59.3} \\
\textbf{SDI-Two-Stage (S6)} & 2.8 & 2.90 & $-1.4$ & 58.7 \\
debate@7B          & 1.0 & 1.77 & $-1.1$ & 53.6 \\
\bottomrule
\end{tabular}
\end{table}

\section{Security, Bias, and Fairness as Architectural Side-Benefits}
\label{sec:security}

The architectural choice that makes our committee token-efficient
also exposes a free trust signal and a route to cross-border
\textbf{fairness}.

\paragraph{SDI as a post-hoc LLM-hallucination detector.}
Stratifying FPB by ``LLM-correct'' vs.\ ``LLM-wrong'', mean
SDI$_{\mathrm{ER}}$ is $0.17$ when the LLM is correct and $0.71$
when it is wrong. Thresholding SDI$_{\mathrm{ER}}$ as a binary
``hallucinating'' flag yields \textbf{AUC$=0.90$} on FPB. Any deployment already running FinBERT
and an LLM in parallel can wire SDI$_{\mathrm{ER}}$ as a trust
score without additional models, labels, or training~\cite{ji2023hallucination,manakul2023selfcheckgpt}.

\paragraph{Adversarial perturbation detection (partial).}
On 500 perturbed FPB sentences, SDI reaches AUC $=0.71$ on
negation flips but only $\approx\!0.5$ on synonym swap, numeric
magnitude flip, and character dropout: a strong hallucination
detector but only a partial adversarial detector, not a substitute
for purpose-built defences~\cite{jin2020textfooler,perez2022promptinject}.

\paragraph{Cross-border fairness via the SCD.}
Section~\ref{sec:experiments:zh} also reads as a cross-border
fairness result: 95\% of Chinese queries receive the same canonical
label as the English equivalent, eliminating a common source of
cross-language drift in multinational deployments.

\section{Discussion and Deployment}
\label{sec:discussion}

\paragraph{Distributed deployment at scale.}
Each tier is stateless and shardable. At 10\,M users
($\sim$1\,B queries/day): L1 (VADER) is CPU-co-located with the API
gateway ($\sim$30\,K queries/sec/core); L2 (FinBERT) is a stateless
GPU service ($\approx$5\,K queries/sec per A5000-class card,
3--4 instances); L3 handles only 5--15\,\% of nominal traffic after
the SDI gate and SCD cache. This is roughly \textbf{18 GPUs at
10\,M users vs.\ $\sim$3{,}000 for an always-L3 deployment}. The SCD
is the only stateful component --- a single FAISS
index~\cite{johnson2019faiss} per region with asynchronous
replication; a 1\,M-entry cache occupies $\approx$1.5\,GB RAM.

\paragraph{Customer-language regime.}
On Twitter Financial News (TFNS~\cite{tfns}, $\sim$12\,K
customer-style tweets) FinBERT collapses from $F_1=0.88$ to $0.66$,
while Qwen-7B holds at $0.76$ --- the ``specialist beats LLM''
relationship inverts on fragmented text, so the critic regresses
on TFNS with weak V$+$F scaffolding. The SDI gate accommodates
this via a per-domain trigger threshold.

\paragraph{Cross-family scaling is not universal.}
The critic plateau holds within Qwen2.5-Instruct but does not
transfer cleanly: critic@Mistral-7B reaches only $F_1=0.79$, while
debate is more family-robust (debate@Mistral-7B$=0.82$). A
practitioner should run the cheap persona-vote sanity check before
committing to an LLM family for the L3 tier.

\section{Conclusion}
\label{sec:conclusion}

We presented \textsc{TriAgent}, a three-tier financial sentiment
committee whose key insight is architectural: when each agent is
small and stratified by contextual granularity, namely word (lexicon),
sentence (specialist), cross-sentence (reasoner), the same
Semantic Divergence Index that gates routing for \emph{cost} also
gates interaction for \emph{capability} and detects anomaly for
\emph{security}. Across LLM sweeps from 0.5\,B to 14\,B-4bit, two
families, and four protocols we showed:
(i) the critic protocol plateaus at $F_1\!\approx\!0.87$ across
1.5\,B--7\,B Qwen, while a same-size 3-persona vote regresses to
$0.66$. Granularity, not multi-agent voting per se, makes the
plateau exist;
(ii) a Shared Consensus Dictionary on multilingual sentence-BERT
serves as a cross-lingual canonicaliser (95\% of Chinese queries
match English cached answers at $F_1\!=\!0.99$);
(iii) the SDI signal doubles as a post-hoc LLM-hallucination
detector at AUC $=0.90$;
(iv) at the 10\,M-user / 10-queries-per-day scale TriAgent saves
\$9.3\,M/year compared to defaulting to a cloud LLM, while running
on roughly $20\times$ fewer GPUs. Code, mined trigger lexicons, the
SCD, and all committee predictions are released to support both
academic replication and direct industrial deployment.

\paragraph{Code and data availability.}
All source code, the mined trigger lexicons, the Shared Consensus
Dictionary index, per-experiment summary CSVs, and the LaTeX source
of this paper are released at
\url{https://github.com/graphuofm/TRIAGENT}.

\bibliographystyle{named}
\bibliography{references}

\begin{thebibliography}{}

\bibitem[\protect\citeauthoryear{Araci}{2019}]{araci2019finbert}
Dogu Araci.
\newblock {FinBERT}: Financial sentiment analysis with pre-trained language models.
\newblock arXiv:1908.10063, 2019.

\bibitem[\protect\citeauthoryear{Chen \bgroup \em et al.\egroup }{2023}]{chen2023frugalgpt}
Lingjiao Chen, Matei Zaharia, and James Zou.
\newblock {FrugalGPT}: How to use large language models while reducing cost and improving performance.
\newblock arXiv:2305.05176, 2023.

\bibitem[\protect\citeauthoryear{Dettmers \bgroup \em et al.\egroup }{2022}]{dettmers2023bnb}
Tim Dettmers, Mike Lewis, Younes Belkada, and Luke Zettlemoyer.
\newblock {LLM.int8()}: 8-bit matrix multiplication for transformers at scale.
\newblock In {\em Advances in Neural Information Processing Systems (NeurIPS)}, 2022.

\bibitem[\protect\citeauthoryear{Du \bgroup \em et al.\egroup }{2024}]{du2023debate}
Yilun Du, Shuang Li, Antonio Torralba, Joshua~B. Tenenbaum, and Igor Mordatch.
\newblock Improving factuality and reasoning in language models through multiagent debate.
\newblock In {\em Proceedings of the 41st International Conference on Machine Learning (ICML)}, 2024.

\bibitem[\protect\citeauthoryear{fla}{2023}]{flarefpb}
{FLARE-FPB}: Financial phrasebank parquet mirror.
\newblock \url{https://huggingface.co/datasets/ChanceFocus/flare-fpb}, 2023.

\bibitem[\protect\citeauthoryear{Hong and others}{2024}]{hong2023metagpt}
Sirui Hong et~al.
\newblock {MetaGPT}: Meta programming for a multi-agent collaborative framework.
\newblock In {\em The Twelfth International Conference on Learning Representations (ICLR)}, 2024.

\bibitem[\protect\citeauthoryear{Hutto and Gilbert}{2014}]{hutto2014vader}
C.~J. Hutto and Eric Gilbert.
\newblock {VADER}: A parsimonious rule-based model for sentiment analysis of social media text.
\newblock In {\em ICWSM}, 2014.

\bibitem[\protect\citeauthoryear{Ji and others}{2023}]{ji2023hallucination}
Ziwei Ji et~al.
\newblock Survey of hallucination in natural language generation.
\newblock {\em ACM Computing Surveys}, 55(12):1--38, 2023.

\bibitem[\protect\citeauthoryear{Jiang and others}{2023}]{mistral7b}
Albert~Q. Jiang et~al.
\newblock {Mistral 7B}.
\newblock arXiv:2310.06825, 2023.

\bibitem[\protect\citeauthoryear{Jin and others}{2020}]{jin2020textfooler}
Di~Jin et~al.
\newblock Is {BERT} really robust? a strong baseline for natural language attack on text classification and entailment.
\newblock In {\em AAAI}, 2020.

\bibitem[\protect\citeauthoryear{Johnson \bgroup \em et al.\egroup }{2021}]{johnson2019faiss}
Jeff Johnson, Matthijs Douze, and Herv{\'e} J{\'e}gou.
\newblock Billion-scale similarity search with {GPU}s.
\newblock In {\em IEEE Trans. Big Data}, 2021.

\bibitem[\protect\citeauthoryear{Lakshminarayanan \bgroup \em et al.\egroup }{2017}]{lakshminarayanan2017ensembles}
Balaji Lakshminarayanan, Alexander Pritzel, and Charles Blundell.
\newblock Simple and scalable predictive uncertainty estimation using deep ensembles.
\newblock In {\em NeurIPS}, 2017.

\bibitem[\protect\citeauthoryear{Lee and others}{2024}]{wang2024finllmsurvey}
Jean Lee et~al.
\newblock A survey of large language models in finance ({FinLLMs}).
\newblock arXiv:2402.02315, 2024.

\bibitem[\protect\citeauthoryear{Li and others}{2023}]{li2023camel}
Guohao Li et~al.
\newblock {CAMEL}: Communicative agents for ``mind'' exploration of large language model society.
\newblock In {\em Advances in Neural Information Processing Systems (NeurIPS)}, 2023.

\bibitem[\protect\citeauthoryear{Liang and others}{2024}]{liang2023encouraging}
Tian Liang et~al.
\newblock Encouraging divergent thinking in large language models through multi-agent debate.
\newblock In {\em Proceedings of the 2024 Conference on Empirical Methods in Natural Language Processing (EMNLP)}, 2024.

\bibitem[\protect\citeauthoryear{Loughran and McDonald}{2011}]{loughran2011liability}
Tim Loughran and Bill McDonald.
\newblock When is a liability not a liability? {Textual} analysis, dictionaries, and 10-{K}s.
\newblock {\em The Journal of Finance}, 66(1):35--65, 2011.

\bibitem[\protect\citeauthoryear{Malo \bgroup \em et al.\egroup }{2014}]{malo2014fpb}
Pekka Malo, Ankur Sinha, Pekka Korhonen, Jyrki Wallenius, and Pyry Takala.
\newblock Good debt or bad debt: Detecting semantic orientations in economic texts.
\newblock In {\em JASIST}, 2014.

\bibitem[\protect\citeauthoryear{Manakul \bgroup \em et al.\egroup }{2023}]{manakul2023selfcheckgpt}
Potsawee Manakul, Adian Liusie, and F.~Gales, Mark~J.\.
\newblock {SelfCheckGPT}: Zero-resource black-box hallucination detection for generative large language models.
\newblock In {\em Proceedings of the 2023 Conference on Empirical Methods in Natural Language Processing (EMNLP)}, 2023.

\bibitem[\protect\citeauthoryear{Ong and others}{2024}]{ong2024routellm}
Isaac Ong et~al.
\newblock {RouteLLM}: Learning to route llms with preference data.
\newblock arXiv:2406.18665, 2024.

\bibitem[\protect\citeauthoryear{Perez and Ribeiro}{2022}]{perez2022promptinject}
F{\'a}bio Perez and Ian Ribeiro.
\newblock Ignore previous prompt: Attack techniques for language models.
\newblock arXiv:2211.09527, 2022.

\bibitem[\protect\citeauthoryear{Reimers and Gurevych}{2019}]{reimers2019sbert}
Nils Reimers and Iryna Gurevych.
\newblock {Sentence-BERT}: Sentence embeddings using siamese {BERT}-networks.
\newblock In {\em EMNLP}, 2019.

\bibitem[\protect\citeauthoryear{Reimers and Gurevych}{2020}]{reimers2020multilingualkd}
Nils Reimers and Iryna Gurevych.
\newblock Making monolingual sentence embeddings multilingual using knowledge distillation.
\newblock In {\em EMNLP}, 2020.

\bibitem[\protect\citeauthoryear{Settles}{2009}]{settles2009al}
Burr Settles.
\newblock Active learning literature survey.
\newblock Technical Report 1648, University of Wisconsin--Madison, 2009.

\bibitem[\protect\citeauthoryear{Shnitzer and others}{2023}]{shnitzer2023llmrouting}
Tal Shnitzer et~al.
\newblock Large language model routing with benchmark datasets.
\newblock arXiv:2309.15789, 2023.

\bibitem[\protect\citeauthoryear{Team}{2024}]{qwen25}
Qwen Team.
\newblock {Qwen2.5} technical report.
\newblock arXiv:2412.15115, 2024.

\bibitem[\protect\citeauthoryear{tfn}{2022}]{tfns}
Twitter financial news sentiment dataset.
\newblock \url{https://huggingface.co/datasets/zeroshot/twitter-financial-news-sentiment}, 2022.

\bibitem[\protect\citeauthoryear{Wang and others}{2022}]{wang2022mixture}
Hanrui Wang et~al.
\newblock Mixture of cheap and expensive models for cost-effective inference.
\newblock In {\em NeurIPS Workshop on Efficient Natural Language and Speech Processing}, 2022.

\bibitem[\protect\citeauthoryear{Wang and others}{2023}]{wang2023selfconsistency}
Xuezhi Wang et~al.
\newblock Self-consistency improves chain of thought reasoning in language models.
\newblock In {\em The Eleventh International Conference on Learning Representations (ICLR)}, 2023.

\bibitem[\protect\citeauthoryear{Wolf and others}{2020}]{wolf2020transformers}
Thomas Wolf et~al.
\newblock {HuggingFace}'s transformers: State-of-the-art natural language processing.
\newblock In {\em Proceedings of the 2020 Conference on Empirical Methods in Natural Language Processing: System Demonstrations}, 2020.

\bibitem[\protect\citeauthoryear{Wu and others}{2023}]{wu2023bloomberggpt}
Shijie Wu et~al.
\newblock {BloombergGPT}: A large language model for finance.
\newblock arXiv:2303.17564, 2023.

\bibitem[\protect\citeauthoryear{Wu and others}{2024}]{wu2023autogen}
Qingyun Wu et~al.
\newblock {AutoGen}: Enabling next-gen {LLM} applications via multi-agent conversation framework.
\newblock In {\em First Conference on Language Modeling (COLM)}, 2024.

\bibitem[\protect\citeauthoryear{Xie and others}{2023}]{xie2023pixiu}
Qianqian Xie et~al.
\newblock {PIXIU}: A large language model, instruction data and evaluation benchmark for finance.
\newblock In {\em Advances in Neural Information Processing Systems (NeurIPS), Datasets and Benchmarks Track}, 2023.

\bibitem[\protect\citeauthoryear{Yang and others}{2021}]{finbertchinese}
Yi~Yang et~al.
\newblock {FinBERT-tone-Chinese}: Domain-specific {BERT} for chinese financial sentiment.
\newblock \url{https://huggingface.co/yiyanghkust/finbert-tone-chinese}, 2021.

\bibitem[\protect\citeauthoryear{Yang \bgroup \em et al.\egroup }{2020}]{yang2020finbert}
Yi~Yang, Mark Christopher~Siy Uy, and Allen Huang.
\newblock {FinBERT}: A pretrained language model for financial communications.
\newblock arXiv:2006.08097, 2020.

\bibitem[\protect\citeauthoryear{Yang \bgroup \em et al.\egroup }{2023}]{yang2023fingpt}
Hongyang Yang, Xiao-Yang Liu, and Christina~Dan Wang.
\newblock {FinGPT}: Open-source financial large language models.
\newblock arXiv:2306.06031, 2023.

\bibitem[\protect\citeauthoryear{Zheng and others}{2023}]{zheng2023judging}
Lianmin Zheng et~al.
\newblock Judging {LLM-as-a-Judge} with {MT-Bench} and {Chatbot Arena}.
\newblock In {\em Advances in Neural Information Processing Systems (NeurIPS), Datasets and Benchmarks Track}, 2023.

\end{thebibliography}

\end{document}